# Aided design of bridge aesthetics based on Stable Diffusion fine-tuning

Leye Zhang[1], Xiangxiang Tian[1,*], Chengli Zhang[2], Hongjun Zhang[3]

(1. Jiangsu College of Finance & Accounting, Lianyungang, 222061, China;

2. Lianyungang Open University, Lianyungang, 222006, China;

3. Wanshi Antecedence Digital Intelligence Traffic Technology Co., Ltd, Nanjing, 210016, China)

**Abstract:** Stable Diffusion fine-tuning technique is tried to assist bridge-type innovation. The bridge real photo dataset is built, and Stable Diffusion is fine tuned by using four methods that are Textual Inversion, Dreambooth, Hypernetwork and Lora. All of them can capture the main characteristics of dataset images and realize the personalized customization of Stable Diffusion. Through fine-tuning, Stable Diffusion is not only a drawing tool, but also has the designer's innovative thinking ability. The fine tuned model can generate a large number of innovative new bridge types, which can provide rich inspiration for human designers. The result shows that this technology can be used as an engine of creativity and a power multiplier for human designers.

## 0  Introduction

In addition to traffic functions, bridges sometimes have landscape value. There are two main ways to express the beauty of bridges: ① The form of structural mechanics and the coordination between the structure and the bridge site environment, which are the basis of bridge aesthetics; ② Appearance decoration, i.e. painting, night scene lighting, additional decorative structures, etc. are used to enhance the aesthetic feeling of the bridge.

In the past 10 years, the technology of generative artificial intelligence has made amazing progress, which provides a new means for designers. Linhao Yang applied Stable Diffusion and its plug-in ControlNet to generate the bridge effect drawing [1]; Yiran Wen applied Stable Diffusion to shape design of automotive parts [2]; Tianchen Zhou applied Stable Diffusion to furniture design [3]; Hongjun Zhang tried to generate new bridge types from latent space of variational autoencoder, generative adversarial network and so on [4-9]; Jiaxin Zhang applied Stable Diffusion to urban renewal of historical blocks [10]; Miao Liu explored the application of Stable Diffusion in the field of industrial design [11]. However, the application of Stable Diffusion model fine-tuning technology in bridge-type design has not been reported.

This paper assumes a landscape bridge-type design task: the owner appreciates The Coral Bridge in Qingdao(the photo is shown in Fig.1). It is hoped that the designer can refer to the shape characteristics of this bridge and design a new bridge-type with similar style but different content. (Note: this time, the appearance decoration method is used to complete the landscape bridge-type design, and the additional decorative structure does not bear the bridge load.)

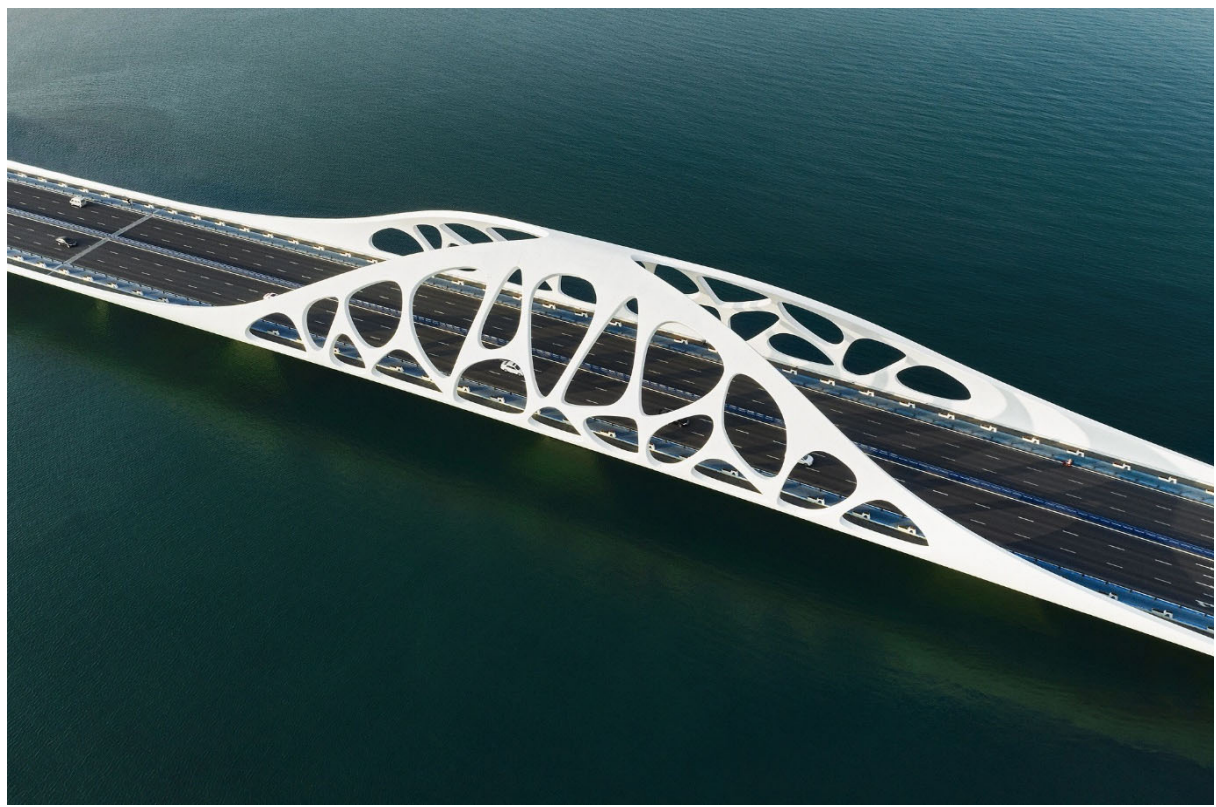

Fig.1 A photo of The Coral Bridge

For this task, if we use the conventional methods of text-to-image, image-to-image, ComfyUI, ControlNet plug-in and so on of Stable Diffusion, it will be very difficult to achieve the goal. Because these conventional methods only regard Stable Diffusion as a draftsman, the task of bridge shape conception is still completed by human designers, which is time-consuming and labor-consuming, and the result may not be satisfactory.

Here, fine-tuning technology is used to complete the task. This method makes Stable Diffusion not only a draftsman but also a designer, replacing human thinking. Taking the real pictures of the bridge in Figure 1 as the dataset, the Stable Diffusion pre training model is fine-tuned by using four methods that are Textual Inversion, Dreambooth, Hypernetwork and Lora. These fine-tuning methods can inject the geometric modeling concept of the dataset into the model, and then use the newly injected concept and the prior knowledge of the model to automatically generate thousands of new bridge types with similar style to the dataset but different geometric shapes within a few hours for the owner and designer to choose (open source address of this article's dataset and source code: https://github.com/ zhangleye/Bridge-SD).

# 1 Introduction to Stable Diffusion
## 1.1 Brief history of text-to-image

Text-to-image is a technique to generate corresponding pictures through some descriptive words. Its development process has four stages: generative adversarial network models, autoregressive models, diffusion models, and diffusion models based on transformers. See the following figure for details:

| GAN-based | Autoregressive models | Diffusion models | Diffusion Transformers |
|---|---|---|---|
| • 2016: GAN-INT-CLS<br>• 2017: AttnGAN<br>• 2019: DM-GAN<br>• 2021: XMC-GAN | • 2020.11: VQ-GAN<br>• 2021.01: DALL·E<br>• 2021.05: CogView<br>• 2022.06: Parti<br>• 2022.07: NUWA-Infinity | • 2021.11: Latent diffusion model<br>• 2022.04: DALL·E 2, Midjourney<br>• 2022.05: Imagen<br>• 2022.05: Make-A-Scene<br>• 2022.08: Stable Diffusion<br>• 2022.11: eDiffi | • 2022.09: UViT<br>• 2022.12: DiT<br>• 2024.01: SiT<br>• 2024.02: Stable diffusion 3 |

Fig.2 The timeline for the development of text-to-image

Stable Diffusion released in 2022 is a completely open source model (the code, training data, and pre training model weights are all open source), and its parameters are small. Most people can inference or even fine tune on ordinary graphics card. Open source of Stable Diffusion has greatly promoted the popularity and development of AIGC (Artificial Intelligence Generated Content). At present, Stable Diffusion text-to-video technology has also made great progress.

## 1.2 Principle of Stable Diffusion

The main structure of the model is composed of three components: VAE (Variational AutoEncoder), Text Encoder (clip model), and Conditional Diffusion model (main component is UNET noise predictor). UNET is operated in the low-resolution space of the image to reduce the need for computing power. VAE is used to compress the input high-resolution image to low-resolution, and decode the result of UNET into a high-resolution output image. Text Encoder converts the input text vector into a context aware sequence and sends it to UNET as image generation condition. UNET generates low-resolution image under the guidance of text condition. Architecture diagram is shown as follows:

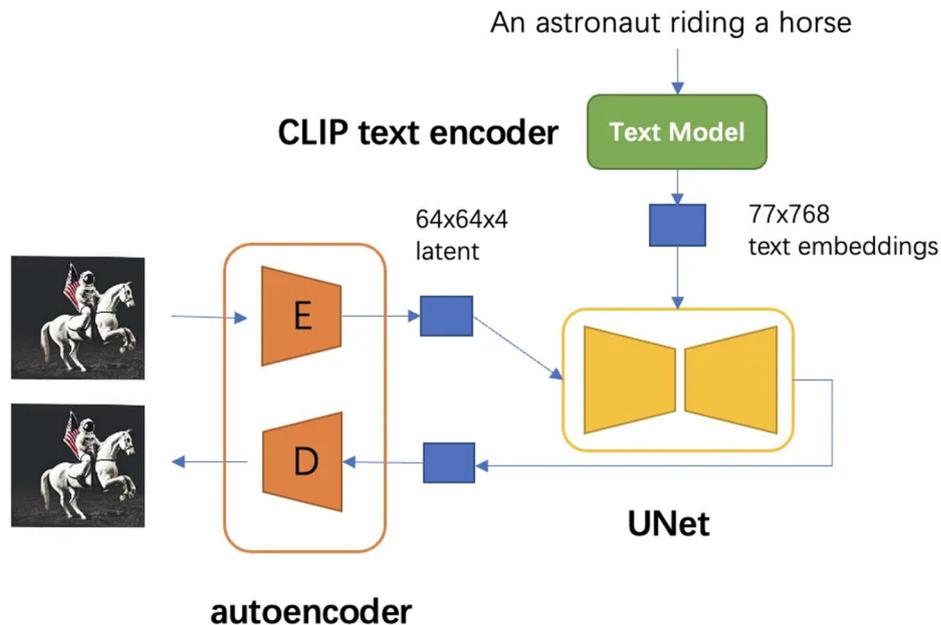

Fig.3 Architecture diagram of Stable Diffusion

During model pre training, only the weight of Conditional Diffusion model is adjusted.

## 1.3 Fine-tuning methods of Stable Diffusion

The model can be personalized through fine-tuning. Fine-tuning methods include Textual Inversion, Dreambooth, Hypernetwork and Lora, etc.

1. Textual Inversion: in Text Encoder, add a new word to represent the personalized content. Adjust the vector of this word in embedding layer through training, so that the word matches the concept in the dataset images.

2. Dreambooth: in the vocabulary of Text Encoder, select rarely used vocabulary to represent personalized content, and adjust the weight parameters such as Conditional Diffusion model through training to make the word match the concept in the images of the dataset. At the same time, the corresponding category word is synchronously trained to retain the prior knowledge of the pre training model.

3. Hypernetwork: a new neural network connected in series with UNET CrossAttention is set up, and its weight parameters are adjusted through training, so that the text description and image of the dataset match each other.

4. Lora: a new neural network in parallel with UNET CrossAttention is set up, and its weight parameters are adjusted through training, so that the text description and image of the data set match each other.

## 2 Software and hardware environments, and self-built dataset

### 2.1 Software and hardware environments

The operating system is WSL Ubuntu 22.04, and the related software versions are Python 3.11.0, TensorFlow 2.15.0, Keras 2.15.0, Keras CV 0.6.0, CUDA version 12.3, Stable Diffusion Webui 1.10.1, Lora scripts-v1.8.5. The CPU is AMD ryzen 5 2600, the memory is 64G DDR4, and the graphics card is RTX4060Ti 16g. Cloud computing is not used.

### 2.2 Self-built dataset

The Coral Bridge in Qingdao is a very creative landscape bridge. Many tourists have shared their photos and videos on the Internet. Here, 20 real bridge photos are collected on the Internet and made into a dataset. The photos are in the format of 512x512 pixels and PNG. Each photo corresponds to a label text file (image-text pair) with the same name. See the following figure for an example:

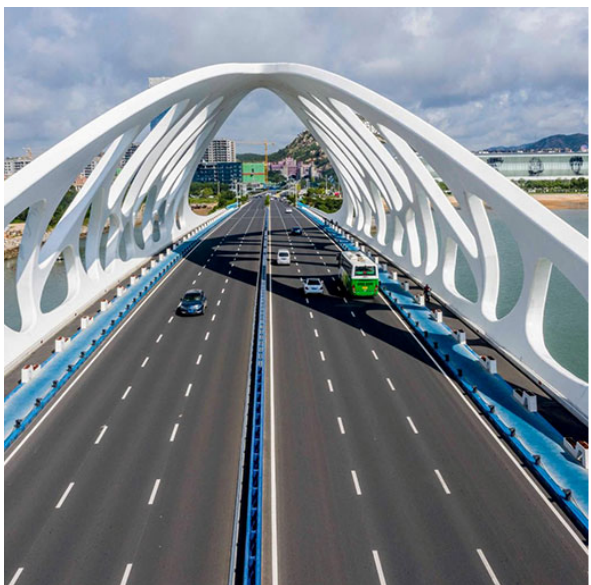

Fig.4 image-text pair dataset

# 3 Aided design of bridge aesthetics based on Stable Diffusion fine-tuning

## 3.1 Fine-tuning method 1: Textual Inversion

(1) Loading dataset and model building

Python 3.11.0, TensorFlow 2.15.0, Keras 2.15.0, Keras CV 0.6.0 are used to build the model. The pre training model is "stable-diffusion-v1-4".

In Text Encoder (existing word serial numbers 0~49407), a new word "<the core bridge>" (word serial number 49408) is added to represent personalized content, and the initial value of its word vector is "bridge".

Load the dataset images, use the general description prompt words (such as "a photo of a<the core bridge>"), and combine them into the image-text pair dataset.

Build the model. See the following figure for the model code:

```python
class StableDiffusionFineTuner(keras.Model):
    def __init__(self, stable_diffusion, noise_scheduler, **kwargs):
        super().__init__(**kwargs)
        self.stable_diffusion = stable_diffusion
        self.noise_scheduler = noise_scheduler

    def train_step(self, data):
        images, embeddings = data
        with tf.GradientTape() as tape:
            latents = sample_from_encoder_outputs(training_image_encoder(images))
            latents = latents * 0.18215
            noise = tf.random.normal(tf.shape(latents))
            batch_dim = tf.shape(latents)[0]
            timesteps = tf.random.uniform((batch_dim,),minval=0, maxval=noise_scheduler.train_timesteps, dtype=tf.int64,)
            noisy_latents = self.noise_scheduler.add_noise(latents, noise, timesteps)
            encoder_hidden_state = self.stable_diffusion.text_encoder([embeddings, get_position_ids()])
            timestep_embeddings = tf.map_fn(fn=get_timestep_embedding, elems=timesteps, fn_output_signature=tf.float32,)
            noise_pred = self.stable_diffusion.diffusion_model([noisy_latents, timestep_embeddings, encoder_hidden_state])
            loss = self.compiled_loss(noise_pred, noise)
            loss = tf.reduce_mean(loss, axis=2)
            loss = tf.reduce_mean(loss, axis=1)
            loss = tf.reduce_mean(loss)
        trainable_weights = self.stable_diffusion.text_encoder.trainable_weights
        grads = tape.gradient(loss, trainable_weights)
        index_of_placeholder_token = tf.reshape(tf.where(grads[0].indices == 49408), ())
        condition = grads[0].indices == 49408
        condition = tf.expand_dims(condition, axis=-1)
        grads[0] = tf.IndexedSlices(
            values=tf.where(condition, grads[0].values, 0), indices=grads[0].indices, dense_shape=grads[0].dense_shape,)
        self.optimizer.apply_gradients(zip(grads, trainable_weights))
        return {"loss": loss}
```

Fig.5 Code for model of Textual Inversion

The loss function is keras.losses.MeanSquaredError(reduction="none").

(2) Training

Use the "Adam" optimizer to update the word vector value of the word "<the core bridge>". The loss curve during training is shown in the following figure:

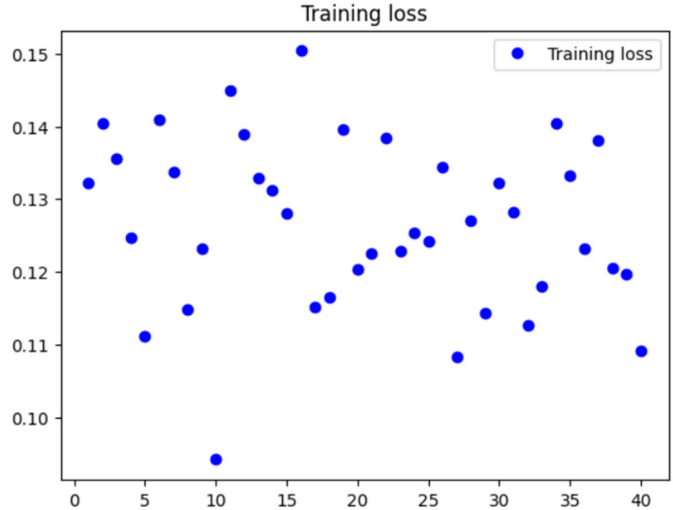

Fig.6 Loss curve of Textual Inversion fine-tuning

(3) Generate new bridge types

Prompt of text-to-image is "a photo of a<the core bridge>", and some new bridge types generated are shown in the following figure: (the results of text-to-image are directly displayed here. If you want to meet the rendering standard, you still need to use the methods of image-to-image and ControlNet plug-in for secondary processing. The same below.)

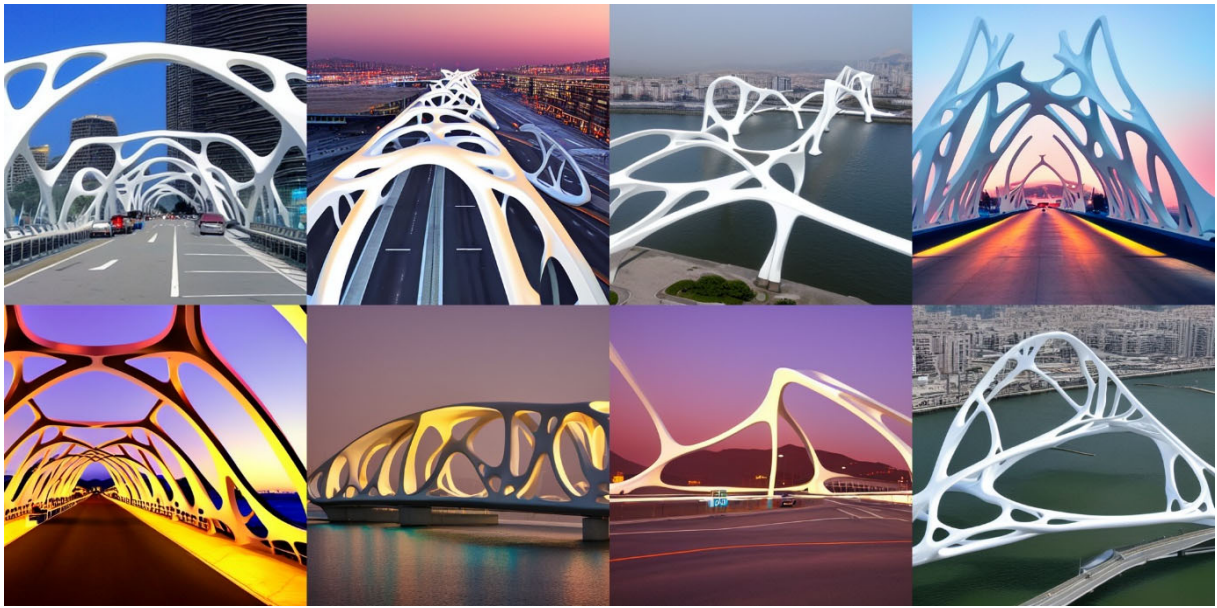

Fig.7 New bridge types generated by Textual Inversion model

## 3.2 Fine-tuning method 2: Dreambooth

Here, Dreambooth plug-in of Stable Diffusion Webui 1.10.1 is used to fine tune the model. The pre training model is "v1-5-pruned-emaonly".

In the Dreambooth plug-in window: create a fine-tuning model, enter instance images directory, prompt of instance images is "a beike bridge", instance token is "beike", and class token is "bridge". The number of class images per instance image is 5, and prompt of class images is "a bridge". Prompt of sampling images is "a beike bridge". Other parameters take default values.

After 4000 steps of training, the loss curve is shown in the following figure:

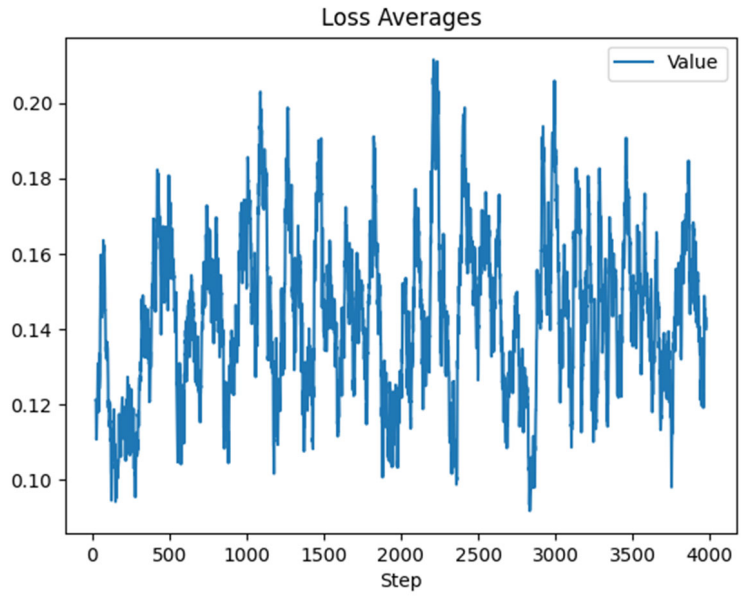

Fig.8 Loss curve of Dreambooth fine-tuning

After the training, Stable Diffusion Webui loads the fine-tuning model. Prompt of text-to-image is "a beike bridge". Some new bridge types are generated as follows:

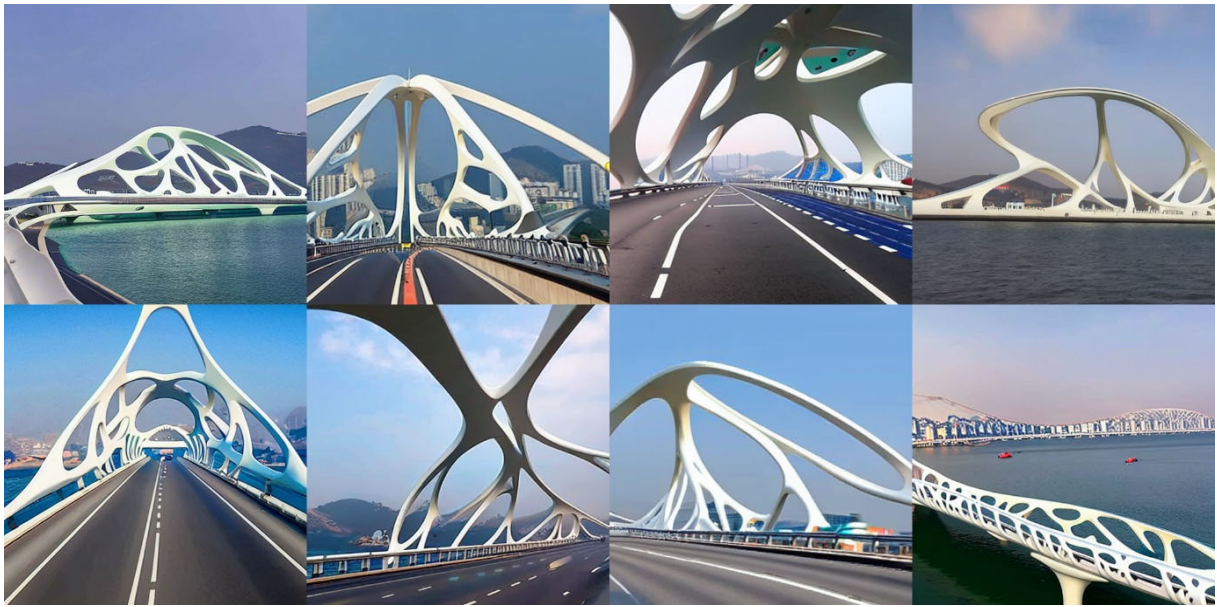

Fig.9 New bridge types generated by Dreambooth model

### 3.3 Fine-tuning method 3: Hypernetwork

Here, Hypernetwork module built into stable diffusion webui 1.10.1 is used to fine tune the model. The pre training model is "v1-5-pruned-emaonly".

In the Hypernetwork module window: create a fine-tuning model, Hypernetwork layer structure parameter is "1,2,1", activation function of Hypernetwork is "linear", and layer weights initialization is "normal". Enter image-text pair dataset directory, and prompt template is "style_filewords.txt" (the content is "a picture of [filewords], art by [name]" and other sentences). Other parameters take default values.

After 1000 steps of training, the loss curve is shown in the following figure:

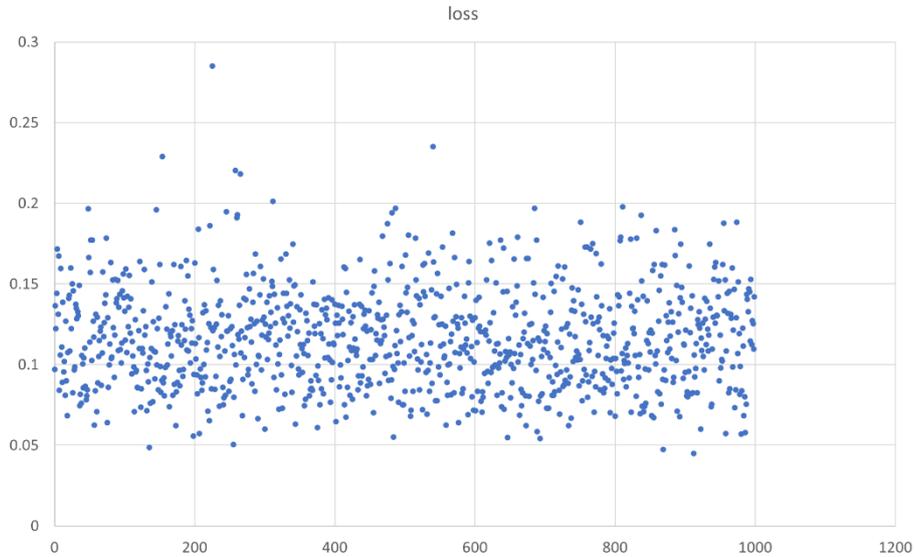

Fig.10 Loss curve of Hypernetwork fine-tuning

After the training, Stable Diffusion Webui adds fine-tuning model to the basic model, and prompt of text-to-image is "a picture of bridge,no humans,outdoors,water,scenery,sky,reflection,day,<hypernet:coral_shell_bridge:1>". Some new bridge types are generated as follows:

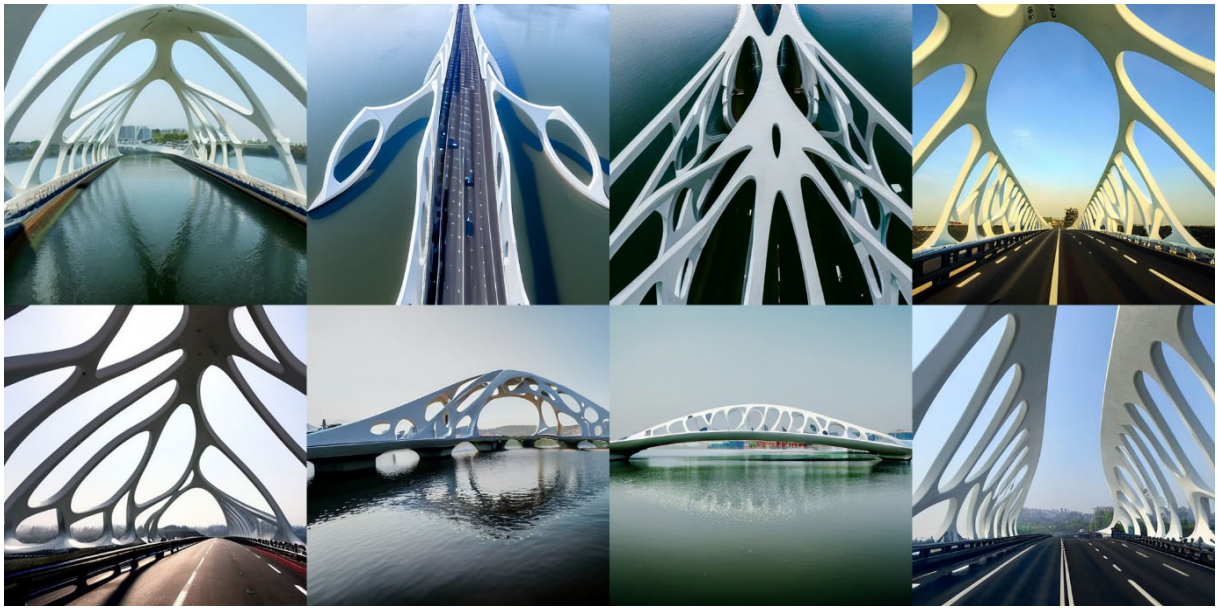

Fig.11 New bridge types generated by Hypernetwork model

### 3.4 Fine-tuning method 4: Lora

Here, lora-scripts-v1.8.5 is used to fine tune the model. The pre training model is "v1-5-pruned-emaonly".

In the Lora module window: create a fine-tuning model, network dimension is 32, and enter image-text pair dataset directory. Other parameters take default values.

After 1000 steps of training, the loss curve is shown in the following figure:

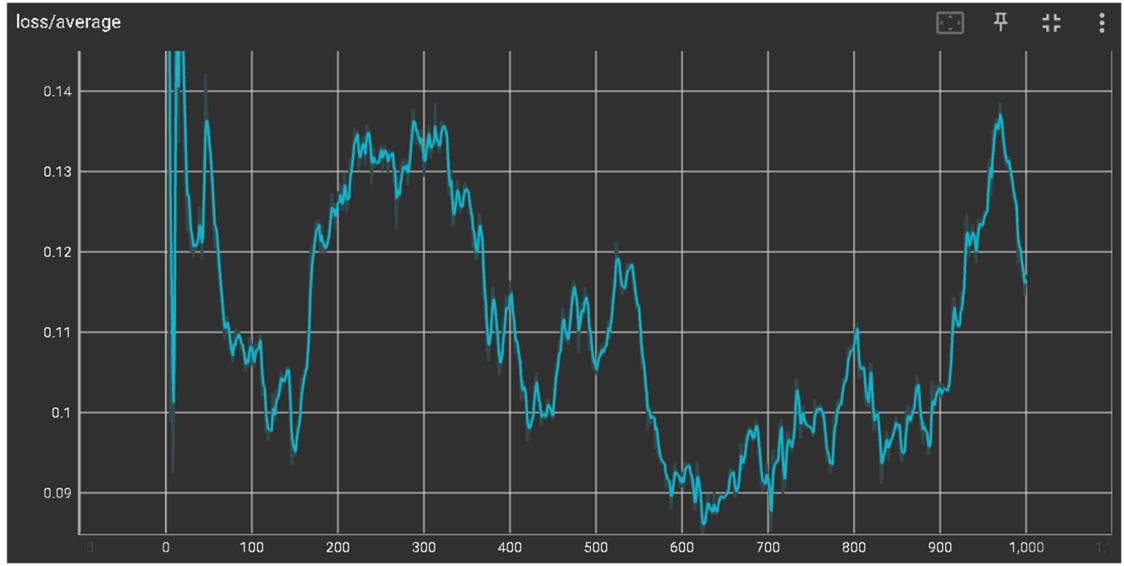

Fig.12 Loss curve of Lora fine-tuning

After the training, Stable Diffusion Webui adds fine-tuning model to the basic model, and prompt of text-to-image is "bridge,no humans,outdoors,water,scenery,sky,reflection,day,<lora:aki:1>". Some new bridge types are generated as follows:

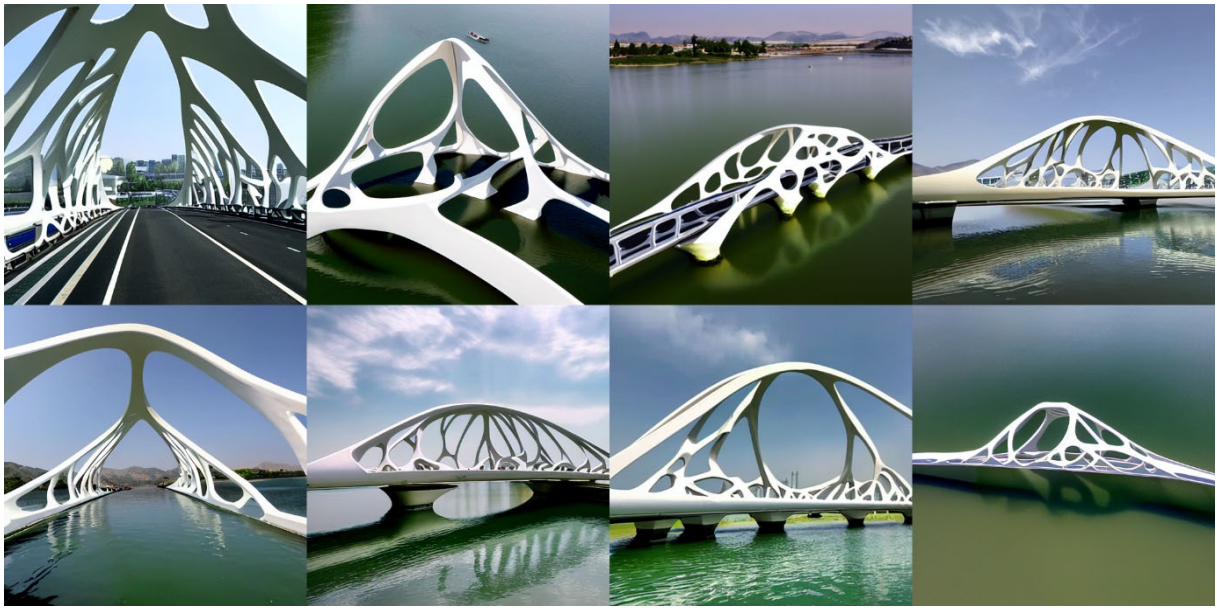

Fig.13 New bridge types generated by Lora model

## 4  Conclusion

(1) The four fine-tuning methods in this paper can capture the main characteristics of the dataset images and realize the personalized customization of Stable Diffusion.

(2) The new bridge types generated by the fine-tuning model is unconstrained and imaginative. It can open up the imagination space and give inspiration to human designers.

(3) Only a small amount of training data is needed to obtain a satisfactory fine-tuning effect. Its efficient learning and generation speed can greatly improve the production efficiency.

(4) The technology used in this paper is also applicable to other design industries (such as industrial design, architectural design, landscape design, etc.).